\documentclass[11pt,a4paper]{article}
\usepackage[hyperref]{eacl2021}
\usepackage{times}
\usepackage{latexsym}

\usepackage{hyperref}
\usepackage{setspace}
\usepackage{float}
\usepackage{amsmath}
\usepackage{amsthm}
\usepackage{amssymb}
\usepackage{booktabs}
\usepackage{adjustbox}
\usepackage{colortbl}
\usepackage{multirow}
\usepackage{enumitem}

\usepackage{microtype}

\aclfinalcopy %

\def\calN{\mathcal{N}}

\def\calL{\mathcal{L}}
\def\R{\mathbb{R}}
\title{Joint Learning of Hyperbolic Label Embeddings\\ for Hierarchical Multi-label Classification}
\newcommand{\model}{\mbox{\textsc{HiddeN}}}
\newcommand{\modeljnt}{\mbox{\textsc{HiddeN}\textsubscript{jnt}}}
\newcommand{\modelcas}{\mbox{\textsc{HiddeN}\textsubscript{cas}}}
\newcommand{\modelflt}{\mbox{\textsc{HiddeN}\textsubscript{flt}}}
\newcommand{\modeleuc}{\mbox{\textsc{HiddeN}\textsubscript{euc}}}
\author{
Soumya Chatterjee\textsuperscript{\rm 1}\thanks{~~Equal contribution} \hspace{0.4em} Ayush Maheshwari\textsuperscript{\rm 1}\footnotemark[1] \\
\textbf{Ganesh Ramakrishnan\textsuperscript{\rm 1} \hspace{0.4em} Saketha Nath Jagaralpudi\textsuperscript{\rm 2}}\\
$^1$ Indian Institute of Technology Bombay\\
$^2$ Indian Institute of Technology Hyderabad \\
\texttt{\{soumya, ayusham, ganesh\}@cse.iitb.ac.in, saketha@iith.ac.in}
}

\date{}

\begin{document}
\maketitle
\begin{abstract}
We consider the problem of multi-label classification, where the labels lie in a hierarchy. However, unlike most existing works in hierarchical multi-label classification, we do not assume that the label-hierarchy is known. Encouraged by the recent success of hyperbolic embeddings in capturing hierarchical relations, we propose to jointly learn the classifier parameters as well as the label embeddings. Such a joint learning is expected to provide a two-fold advantage: i) the classifier generalises better as it leverages the prior knowledge of existence of a hierarchy over the labels, and ii) in addition to the label co-occurrence information, the label-embedding may benefit from the manifold structure of the input datapoints, leading to embeddings that are more faithful to the label hierarchy. We propose a novel formulation for the joint learning and empirically evaluate its efficacy. The results show that the joint learning improves over the baseline that employs label co-occurrence based pre-trained hyperbolic embeddings. Moreover, the proposed classifiers achieve state-of-the-art generalization on standard benchmarks. We also present evaluation of the hyperbolic embeddings obtained by joint learning and show that they represent the hierarchy more accurately than the other alternatives.\footnote{Code can be found at \url{https://github.com/soumyac1999/hyperbolic-label-emb-for-hmc}}
\end{abstract}

\section{Introduction}
The problem of multi-label text classification is well known and extensively studied in literature~\citep{mccallum1999multi,yang2009effective,liu2017deep}. The fundamental assumption is that a document is associated with multiple labels from a fixed vocabulary of labels. Often, these labels are organised in a hierarchical structure. We undertake the task of labelling documents with classes that are hierarchically organised; this problem is popularly known as \textit{hierarchical multi-label text classification} (HMC). HMC methods have found several applications in online advertising systems~\citep{agrawal2013multi}, bio-informatics~\citep{peng2016deepmesh, triguero2016labelling}, text classification~\citep{rousu2006kernel, emnlp}.

The main challenge in HMC is in modelling classification of the document into a large, imbalanced and structured output space. In HMC, the label taxonomy is a partially ordered set ($L, \prec)$ where $L$ is a finite set of all class labels. Relation $\prec$ refers to \textit{is-a} relationship between labels, which is asymmetric, anti-reflexive and transitive~\citep{survey}.

Hierarchical structures can provide important insights for learning and classification tasks. However, explicit knowledge of hierarchy is not available in several domains, for instance, extreme classification datasets~\cite{extreme}. In this paper, we consider the problem of structured prediction from unstructured text, in which label hierarchy is not known \textit{apriori}. We infer hierarchies from classification judgements on the outputs that are readily available. We focus on discovering relationships between the labels in a hyperbolic space, which has natural capacity to encode hierarchical structures.

In our approach, \model\ (\textbf{H}yperbol\textbf{I}c label embe\textbf{DD}ings for hi\textbf{E}rarchical multi-label classificatio\textbf{N}), the labels are represented in a hyperbolic space to help respect their latent hierarchical organisation. We use this intuition to learn label embeddings for HMC without explicit supervision on the label hierarchy.

Apart from employing hyperbolic embeddings, another key aspect of our methodology is that the parameters of the classifier as well as of the label embedding are learnt jointly. We next explain the advantage in doing so. In the absence of any partial information regarding the hierarchy, label embeddings are typically learnt using the weak supervision available in label co-occurrences~\cite{poincare,lorentz}. This weak form of supervision can be complemented if the label embedding learning is also aware of the manifold structure of the input (documents). For {\em e.g.},  similar documents may have similar labels, {\em etc.} Such a strengthening is possible only if learning happens in a joint fashion. Moreover, the generalization of the classifier also improves because of the improved embeddings (and vice-versa).
Our contributions can be summarised as follows:
\begin{enumerate}[topsep=0pt,itemsep=-1ex,partopsep=1ex,parsep=1ex,leftmargin=*]
    \item We present an approach \model, that models the implicit hierarchical organisation of labels for improved classification. It leverages properties of hyperbolic geometry to help learn embeddings for the  hierarchically organised labels.
    \item We present a novel formulation for jointly learning the parameters of the classifier as well as the label embedding, which can be trained solely using the supervision from the training data, and without using any explicit information regarding the label hierarchy. 
    \item We evaluate \model\ on real-world  as well as synthetic datasets and show:
    \begin{enumerate}[topsep=-1ex,itemsep=-1ex,partopsep=1ex,parsep=1ex,leftmargin=1.5em]
        \item significant improvement over classical multi-label classification methods as well as baselines that employ hyperbolic label embeddings learnt in isolation solely based on label co-occurrence information
        \item \model\ sometimes generalizes even better than state-of-the-art hierarchical multi-label classifiers that have complete access to the true label hierarchy
        \item label embeddings learnt using the joint optimisation approach correlate better with the ground truth than other alternatives.
    \end{enumerate}
\end{enumerate}

\section{Related Work}
Several conventional classification methods are capable of handling classification in multi-label settings. However, relatively fewer of these are designed to incorporate the possibly hierarchical organisation of the class labels. These include both traditional methods~\citep{gopal2013recursive, rcv1} as well as deep learning methods~\citep{johnson2015effective,peng2018large} across varied domains such as news articles, web content, \emph{etc}.

Traditional or flat classification approaches typically perform prediction assuming that all the classes are independent of each other, ignoring the class hierarchy. Whereas `local' classification approaches~\citep{koller1997hierarchically, cesa2006hierarchical} train a set of classifiers at each level of the hierarchy. However, it has also been argued~\citep{cerri2011hierarchical} that it is impractical to train separate classifiers at each level. On the other hand, `global' approaches~\citep{silla2009global, wang2001hierarchical} train a single classifier that factors in the complete class hierarchy. Unlike the local approach, `global' approaches do not suffer from the error propagation problem, although they are prone to under-fit by not considering local information in the hierarchy.

Some recent papers have proposed a mix of local and global approaches for HMC. \citet{hmcn} propose an objective that leverages both local and global information while introducing global hierarchical violation penalty. \citet{emnlp} employ a reinforcement learning framework to learn a label assignment policy. They model HMC as a \textit{markov decision process}, wherein, the agent takes an action of label assignment on the tree hierarchy  and receives scalar rewards as feedback for the actions. \citet{hyperim} embed both document and label hierarchy in the same hyperbolic space and use interactions between these embeddings for HMC. Our approach differs from these in two important ways: (i) we embed only labels into the hyperbolic space and (ii) label hierarchy is not known \textit{apriori} - all we assume is that there is some hidden hierarchy. 

Recently, the use of hyperbolic geometry has been found to be promising in machine learning and network sciences to model data with latent hierarchies. \citet{complexhyperbolic} showed that properties of complex networks, namely {\em heterogeneous degree distribution} and {\em strong clustering}, naturally manifest in hyperbolic geometry. They showed that if a network has some heterogeneous degree distribution and metric structure,  the network can be mapped effectively to the hyperbolic space (since euclidean distance has limitations in approximating the distance between nodes in a tree). \citet{hyperbolic1987} have shown that any finite tree structure can be embedded into a finite hyperbolic space while preserving the distance between nodes. \citet{poincare} learnt hierarchical representations of symbolic data by embedding them into an $n$-dimensional Poincar\'e ball by leveraging the distance property of hyperbolic spaces. Instead of relying on the true hierarchy to learn embeddings, \citet{lorentz} inferred hierarchies from real-valued similarity scores using the Lorentz model of hyperbolic geometry. We used a similar formulation in our model \model\ to build our HMC model by leveraging the co-occurrence count of labels for each document, but additionally (and more importantly), in a joint manner, learn the parameters of the classifier.

\section{Hyperbolic Geometry \& the Poincar\'e Model}\label{sec:backpoi}
In this section, we give an overview of hyperbolic geometry and the Poincar\'e model for embedding in hyperbolic spaces~\citep{poincare}.
A hyperbolic space is a non-Euclidean Riemannian manifold of constant negative curvature. Though there are several fundamental differences between the Euclidean and the hyperbolic geometry, the most interesting characteristic of hyperbolic spaces is their ability to naturally represent hierarchical relations \citep{complexhyperbolic}.
In the Poincar\'e ball model, which is one of the standard models for hyperbolic geometry, the Euclidean distances between equidistant points, according to the inherent manifold metric $d$, fall exponentially as one moves from origin towards the surface of the ball. This interesting property is the key for enabling learning of continuous embeddings of hierarchies. For example, one can imagine root node of hierarchy at origin and leaf nodes near the ball's surface. Then, this model can easily accommodate exponentially growing number of equidistant siblings at deeper levels of the hierarchy. Whereas, such an accommodation is not possible using Euclidean geometry. Below we provide some details of this model.\par
Let $\mathcal{B}^{n} = \{x \in \mathbb{R}^{n} \vert \   \|x\| < 1\}$ be the \textit{open n-}dimensional unit ball, where $\|.\|$ is the Euclidean 2 norm. The Poincar\'e ball model is a Riemannian Manifold $(\mathcal{B}^{n}, g_{x})$, the open unit ball equipped with the Riemannian metric tensor $g_{x} = \Bigg(\frac{2}{1-{\|x\|}^2}\Bigg)^{2}g^{E}$, where $x \in \mathcal{B}^{d}$ and $g^{E}$ is the Euclidean metric tensor. The geodesic distance between two points $u, v \in \mathcal{B}^{d}$ is given as
\begin{equation}
   d(u,v) = arcosh\Bigg( 1 + 2\frac{\|u-v\|^{2}}{(1-\|u\|^{2})(1-\|v\|^{2})}\Bigg)
  \label{eq:dpoincare}
\end{equation}
Given any $x\in\R^n$, one can show that $\frac{x}{1+\sqrt{1+\|x\|_2^2}}$ always lies in the Poincare ball~(refer Appendix for detailed explanation).

\section{Problem Formulation and Approach}
In this section, we present present details of our model, training, as well as inference.

\subsection{Problem Formulation \label{sec:probformulation}} 
Here we consider an interesting special case of multi-label classification. The training data is of the form: $\mathcal{D} = \left\{\left(D_1,y_1\right), \left(D_2,y_2\right), \ldots, \left(D_m,y_m\right)\right\}$, where $D_i\in\R^n$ is the input representation of the $i^{th}$ document, $y_i\in\{0,1\}^L$ represents the set of active/annotated labels for it ($y_i^l=1\ \iff\ D_i$ is labelled with $l$), and $L$ is the total number of labels. Importantly, the labels are assumed to be nodes of an unknown, yet fixed, hierarchy.  Using this prior knowledge and the training data, the goal is to learn a classifier that generalises well for labelling new documents.

Classical text classification methods ignore the informative prior knowledge that the set of labels form a hierarchy. Most of the hierarchical multi-class classification models assume that the hierarchy over the labels is completely known, which might not be a pragmatic assumption, since constructing hierarchies is an expensive process, especially when the number of labels is large~\cite{extreme}.  In contrast, here we assume no explicit information regarding the hierarchy other than it's existence, and the implicit information encoded in the training data. Also, in our set-up, we do not restrict the labels to be the leaves nodes in the hierarchy. As motivated earlier, here we propose to learn a classifier that jointly learns the classifier parameters as well as the label embeddings.

\subsection{Our Model: \model}
Our proposed model \model\ has two key components: one for representing the documents that may lead to well-generalizing classifiers and the other for embedding the labels in a hyperbolic space. Recall that hyperbolic spaces have shown to be well-suited for data satisfying hierarchical relations.

\textbf{Document Model} $\mathcal{F}_w$ accepts as input a document, $D$, and outputs a $n$-dimensional representation of it, $\mathcal{F}_w(D)\in\R^n$. Here, $w$ is the set of parameters to be learnt. In this work, we use TextCNN \citep{textcnn} as the document model. But our approach remains valid irrespective of the chosen document model.

\textbf{Label Embedding Model} $\mathcal{G}_\Theta$ accepts as input a label $l$ and outputs  a finite dimensional representation, $\mathcal{G}_\Theta(l)$. Here, $\Theta$ is the set of parameters to be learnt. In this work, following~\citet{lorentz}, we employ the simple look-up based model defined by $\mathcal{G}_\Theta(l)\equiv \Theta*y^l=\Theta_l$, where $\Theta\in\R^{n\times L}$ and $\Theta_l$ is the $l^{th}$ column of $\Theta$. These Euclidean embeddings $\Theta_l$ are then projected onto the Poincare manifold using the transformation $\Pi(x) = \frac{x}{1+\sqrt{1+\|x\|_2^2}}$. In summary, the hyperbolic embedding of label, $l$, is given by $\Pi(\Theta_l)$.

We next assume that there exists some optimal set of parameters $w^*,\Theta^*$ such that the labels annotated/active for a document, $D$, are exactly those whose label representations are highly aligned with that of $D$'s representation. Here, alignment between the representations is intended to model the natural intuition of appropriateness between label and document. Following the principle of large-margin separation, in this paper we employ the \textbf{alignment model} defined below:
\begin{equation}
    \hat{y}_D^l\left(w,\Theta\right) \equiv \sigma\left(\mathcal{F}_w\left(D\right)^\top\Theta_l\right)
\end{equation}
where $\hat{y}_D^l\left(w,\Theta\right)$ denotes the alignment between the document, $D$, and the $l^{th}$ label as per the model with parameters $\left(w,\Theta\right)$, and $\sigma$ is the Sigmoid activation function.\par

\textbf{Inference:} Given the learnt parameters $(\hat{w},\hat{\Theta})$, the labels with $\hat{y}_D^l(\hat{w},\hat{\Theta})>0.5$ are predicted to be the active ones for $D$. We next detail the proposed joint objective for learning the parameters.

\subsection{Joint Objective \label{sec:jointObjective}}
The proposed objective consists of two terms: the first is an empirical multi-label loss term over the training data, and the second is a loss for ensuring that the hyperbolic label embeddings respect the pairwise label co-occurrence or any other such (pairwise) partial information regarding the underlying label hierarchy.

\textbf{First Term} is simply a binary cross entropy loss to promote high alignment scores for each annotated label and vice-versa:
\begin{align}\nonumber
    \mathcal{L}_1\left(w,\Theta\right) &= \sum_{i=1}^{m} \sum_{l=1}^L \left[y_i^l \log\left(\hat{y}_i^l\left(w,\Theta\right)\right)\right.\\
    &+ \left.(1-y_i^l)\log\left(1-\hat{y}_i^l\left(w,\Theta\right)\right)\right]
    \label{eq:comp1}
\end{align}\par
where $\hat{y}_i^l$ is a short-hand for $\hat{y}_{D_i}^l$.

\textbf{Second Term} induces lesser geodesic distance in the hyperbolic space between the label embeddings that have higher co-occurrences than those between label pairs that have less co-occurrence~\cite{lorentz}:
\begin{equation}
    \mathcal{L}_2(\Theta) = \sum_{\substack{l, l^\prime \in L,\\l^{\prime}\ne l}}
    \log\left(\frac{e^{-d\left(\Pi\left(\Theta_l\right), \Pi\left(\Theta_{l^\prime}\right)\right)}}{\sum\limits_{z\in\calN\left(l, l^{\prime}\right)}e^{-d\left(\Pi\left(\Theta_l\right), \Pi\left(\Theta_{l^\prime}\right)\right)}}\right)
    \label{eq:comp2}
\end{equation}
where $d$ is the metric in the hyperbolic space given by Eq.\ref{eq:dpoincare}, $\Pi(\Theta_l)$ is hyperbolic embedding of $l^{th}$ label, and $\calN\left(l, l^{\prime}\right)$ is the set of all labels that less frequently co-occur with $l$ than $l^\prime$ co-occurs with $l$.

The overall objective function is a weighted sum of the two components described above
\begin{equation}
    \calL\left(w,\Theta\right) = \calL_1\left(w,\Theta\right) + \lambda \calL_2\left(\Theta\right)
    \label{eq:objective}
\end{equation}

We refer to the model corresponding to the parameters $(w_{\textup{jnt}},\Theta_{\textup{jnt}})$ that minimizes this joint objective in Eq.\ref{eq:objective} as \textbf{\modeljnt}:
\begin{equation}
    (w_{\textup{jnt}},\Theta_{\textup{jnt}})\in\arg\min_{w,\Theta} \calL(w,\Theta)
\end{equation}

Both components of our loss interact with each other to minimize the distance between document and label embeddings in the hyperbolic space. The advantage of the joint learning is well illustrated when \modeljnt\ is compared with the following baseline, henceforth referred to as \textbf{\modelcas}:
\begin{enumerate}[label=(\arabic*),topsep=0pt,itemsep=-1ex,partopsep=1ex,parsep=1ex,leftmargin=*]
    \item $\calL_2$ is minimized to obtain label embeddings $\hat{\Theta}_{\textup{cas}}\in\arg\min_{\Theta}\calL_2\left(\Theta\right)$.
    \item These are then used in $\calL_1$ to obtain document parameters: $\hat{w}_\textup{cas}\in\arg\min_{w}\calL_1(w,\hat{\Theta}_\textup{cas})$.
\end{enumerate}

We also empirically compare with the following multi-class classification baseline, henceforth referred to as \textbf{\modelflt}:
\begin{enumerate}[label=(\arabic*),topsep=0pt,itemsep=-1ex,partopsep=1ex,parsep=1ex,leftmargin=*]
    \item $\Theta_{\textup{flat}}$ is fixed to the identity matrix.
    \item These are then used in $\calL_1$ to obtain document parameters: $\hat{w}_\textup{flat}\in\arg\min_{w}\calL_1(w,\hat{\Theta}_\textup{flat})$. 
\end{enumerate}

To evaluate the benefit of using hyperbolic spaces for embedding labels, we also compare with a variant of \modeljnt\ called \textbf{\modeleuc}\ for which $\calL_2$ is modified to be
\begin{equation}
    \calL_{2\text{Euc}}(\Theta) = \sum_{\substack{l, l^\prime \in L,\\l^{\prime}\ne l}}
    \log\left(\frac{e^{-\left\|\Theta_l - \Theta_{l^\prime}\right\|_2}}{\sum\limits_{z\in\calN\left(l, l^{\prime}\right)}e^{-\left\|\Theta_l - \Theta_{l^\prime}\right\|_2}}\right)
    \label{eq:comp2euc}
\end{equation}
to obtain the parameters $(w_{\textup{euc}},\Theta_{\textup{euc}})$ as $(w_{\textup{euc}},\Theta_{\textup{euc}})\in\underset{w,\Theta}{\arg\min}\ \calL_1(w,\Theta) + \lambda \calL_{2\text{Euc}}(\Theta)$

Note that none of the variants of \model\ assume any explicit information regarding the underlying hierarchy. However, the former three exploit the prior knowledge that there exists a label hierarchy; whereas the latter, which is the classical multi-label classification network, completely ignores this useful information. Moreover, since the proposed model \modeljnt\ performs joint learning, it is expected that \modeljnt\ not only achieves better generalization, but also leads to better label embeddings, when compared to \modelcas. The simulation results in section~\ref{sec:sim} confirm the same.

\subsection{Training Details}
In all our experiments, the initial word embedding layer of TextCNN in the document model is initialized using 300 dimensional GloVe embeddings~\citep{glove}. Following \citet{poincare}, we randomly initialize $\Theta$ from the uniform distribution $\mathcal{U}(-0.001, 0.001)$. Both the document and label representations are are of length $n=300$. We randomly choose 10\% of training set as the validation set and report test set results on the best validation epoch. During training, dropout is applied to the outputs of document model as well the label model with probabilities $0.1$ and $0.6$ respectively. We found $\lambda=0.1$ to yield the best validation performance. The number of training epochs are set to $30$ for all experiments. Both models are optimized using stochastic gradient descent using Adam optimizer \citep{kingma2014adam} with learning rate as $0.001$ for TextCNN.

We run all our experiments on Nvidia RTX 2080 Ti GPUs 12 GB RAM over Intel Xeon Gold 5120 CPU having 56 cores and 256 GB RAM. It takes around 1, 2 and 5 hours to train the model on RCV1, NYT and Yelp datasets respectively.

\section{Experiments}\label{sec:sim}
In this section, we compare our approach against the baseline models and other state-of-the-art HMC approaches. First we describe the evaluation metrics and illustrating results in a synthetic setting.

\subsection{Evaluation Measures}

\textbf{Classifier Evaluation Measures}: We use standard measures for evaluating any HMC, {\em viz.}, \textbf{Macro-F1} and \textbf{Micro-F1}. Let $TP$, $TN$, $FN$, $FP$ denote the true positive, true negative, false negative and false positive labels respectively. Precision is $\frac{TP}{TP+FP}$ and recall is $\frac{TP}{TP+FN}$. F1-score is the harmonic mean of precision and recall. \textbf{Macro-F1} assigns equal weightage to each class and is computed as the averaged F1 score over all classes. \textbf{Micro-F1} is the F1-score computed over all instances.

\noindent \textbf{Label Embedding Evaluation Measures}:
For a given application, let us say $\mathcal{H}_{*}$ is provided to us as the ground truth hierarchy of labels/nodes, which was assumed to be unknown in our problem formulation in Section~\ref{sec:probformulation}. Recall that none of the variants of \model\ has access to $\mathcal{H}_{*}$. How consistent with respect to $\mathcal{H}_{*}$ are the label embeddings learnt by these models? We attempt to assess this consistency by adopting standard measures such as Spearman's rank correlation coefficient~\citep{zar2005spearman} and Normalized Discounted cumulative gain (NDCG)~\cite{dcg}. 

Recall from Section~\ref{sec:jointObjective}, that the hyperbolic embedding for label $l$ is $\Pi\left(\Theta_l\right)$ and likewise for $l'$, it is $\Pi\left(\Theta_{l^\prime}\right)$. The model parameters $\Theta$ might be learnt using any variant of \model. Given a query label, $l$, the geodesic distance $d(\Pi\left(\Theta_{l}\right),\Pi\left(\Theta_{l^\prime}\right))$ is used to rank all other labels $l'\ne l$; smaller the distance, larger the rank. Any two labels $l'\ne l''$ that are at the same geodesic distance from $l$, will be assigned the same rank $r = r''$. Next, we define a graded relevance score for labels with respect to the ground truth hierarchy, $\mathcal{H}_{*}$. 
For any given query label $l \in \mathcal{H}_{*}$, we also assign a graded relevance $rel' \in \mathbb{N}$ to every other label $l' \ne l$  based on the distance (number of hops $hops(l,l')$) of $l'$ from $l$ in the hierarchy $\mathcal{H}_\mathcal{D}$; smaller the distance, larger the graded relevance (we considered $rel' \propto \frac{1}{hops(l,l')}$, for example).

\textit{Discounted Cumulative Gain (DCG)}~\citep{dcg} is a standard measure of the quality of ranking of an approach with respect to the graded relevance provided in the ground truth. DCG@k measures items ({\em eg:} labels $l'$) $k$ hops away from the query $l$. The gain is accumulated from the top of the ranked list upto some pre-specified position $k$ in the list, with the gain of each result discounted at lower ranks: $DCG_k = \sum_{i=1}^k \frac{rel_i}{log_2(i+1)}$. Here, $rel_i$ is the graded relevance at position $i$. This result is itself averaged over all query labels $l \in \mathcal{H}_{*}$. 

\textit{Spearman's rank correlation coefficient} denoted by $r$ is a non-parametric metric to measure statistical dependence between ranking of two variables. For each query label ($l$), we first measure rank correlation between the predicted rank $r_p$ of a label $l'$ and its own rank $r_h$ as per ground truth hierarchy $h$. The correlation coefficient between $r_{l_p}$ and $r_{l_h}$ is computed as $r_l = \frac{cov(r_{p}, r_{h})}{\sigma_{r_{p}}~\sigma_{r_{h}}}$. Here, $cov$ is the covariance and $\sigma$, the standard deviation. The final score, $r$ is the averaged score across all labels $l'$ in the set.

\begin{figure}
    \centering
    \includegraphics[width=.3\textwidth]{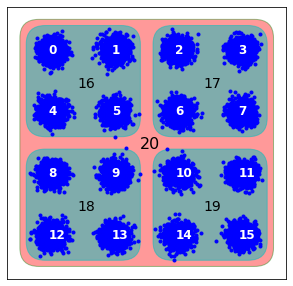}
    \caption{Gaussian used for the synthetic experiment}
    \label{fig:synthetic}
\end{figure}

\subsection{Validation via Synthetic Experiments}
To observe the behaviour of the proposed approach \model\ with respect to the evaluation measures in a controlled environment, we present one such synthetic setup. The goal in this section is to illustrate the advantage of joint learning of parameters over isolated learning. Consider 2D data generated from 16 neatly separated Gaussians laid out on a grid as illustrated in Figure~\ref{fig:synthetic}. Each of the 16 gaussians corresponds to a single label $l_1, l_2... l_{16}$. We consider a second layer of 4 labels $l_{17}...l_{20}$, obtained by grouping the gaussians into 4; each quadrant in the larger square would correspond to a single label.  Finally, we have a third layer consisting of a single label $l_{21}$ - {\em viz.}, the entire large square. This simple hierarchy is hidden from our variants of \model\ as well as from the flat model. The synthesized data is split randomly into train-test in the ratio 60:40. For each of the jointly optimised model \modeljnt, the cascaded model \modelcas\ as well as the flat model, we observe (i) the performance of the classification  models $\mathcal{F}_w(D)$ measured in terms of \textbf{Micro-F1} and \textbf{Macro-F1} as well as (ii) the consistency of the label embedding models $\mathcal{G}_\Theta(l)$ with respect to the hidden 3-level hierarchy over the $21$ labels. We record observations in two settings:

{\bf \mbox{\textsc{Setting 1}}} in which {\em for each training instance, one of the annotated labels are dropped, uniformly at random:} In Table~\ref{tab:syn1} we note the performances of the different approaches with increasing rate at which labels are dropped. The jointly optimised model \modeljnt\ accounts for the classification task through loss component ${\mathcal L}_1$ as also the somewhat redundant label co-occurrence through the loss component ${\mathcal L}_2$. 

As expected, we observe that the performance of \modeljnt\ is more robust to this form of label noise than the \modelcas\ and \modelflt models. This is because \modelflt\ entirely relies on the training data and ignores the prior knowledge of existence of a label hierarchy. \modelcas\ is also less robust as it over-relies on the label co-occurrence by minimising ${\mathcal L}_2$ (which, in isolation will be sensitive to label noise), before venturing into the classification task by minimising ${\mathcal L}_1$.

\begin{table}
\caption{Comparison of methods on the Synthetic {\bf \mbox{\textsc{Setting 1}}} 
with respect to increasing probability with which a label is randomly dropped. Synthetic data used here has 12000 training and 8000 test samples.}
\label{tab:syn1}
\begin{adjustbox}{max width=0.48\textwidth}
\begin{tabular}{l|cc|cc|cc}
\hline
\multirow{2}{*}{Prob} & \multicolumn{2}{c|}{0.00} & \multicolumn{2}{c|}{0.20} & \multicolumn{2}{c}{0.40} \\ \cline{2-7} 
 & Micro F1 & Macro F1 & Micro F1 & Macro F1 & Micro F1 & Macro F1 \\ \hline
\modelflt & 96.8 & 89.1 & 93.2 & 87.8 & 90.4 & 87.7 \\
\modelcas & 98.0 & 93.4 & 94.4 & 88.9 & 91.9 & 91.0 \\
\modeljnt & 98.1 & 94.0 & 94.8 & 91.6 & 92.3 & 91.7 \\
\hline
\end{tabular}
\label{tab:synth1}
\end{adjustbox}
\end{table}

{\bf \mbox{\textsc{Setting 2}}} in which the size of the {\em training set is decreased without corrupting labels:} 
We observe the performances of the different approaches with decreasing size of the training set and note that the performance of the jointly optimised model \modeljnt\ falls back on the label correlation signals through the loss component ${\mathcal L}_2$ and is therefore more robust to decreasing size of the data set than the flat classifier. We observed similar results for other synthetic settings. Owing to space constraints, the plots and other ranking results are provided in the supplementary material.

\subsection{Real-world Text Datasets}
We used three datasets, namely, RCV1, Yelp and NYT in our experiments:
\begin{enumerate}[label=(\arabic*),topsep=0pt,itemsep=-1ex,partopsep=1ex,parsep=1ex,leftmargin=*]
\item \textbf{RCV1} \citep{rcv1} - RCV1 is a newswire dataset of the articles collected between 1996-1997 from Reuters
\item \textbf{NYT} \citep{nytimes} - This corpus contains articles from New York Times published between January 1st, 1987 and June 19th, 2007
\item \textbf{Yelp} \footnote{\url{https://www.yelp.com/dataset/challenge}} - Yelp is a review dataset of restaurants and each review is labelled with hierarchical categories of restaurants. Following the experimental design in \citet{emnlp}, we use the set of reviews for a business to predict the categories to which the business belongs.
\end{enumerate}

Some statistics pertaining to these datasets are presented in Table \ref{tab:data}.

\begin{table}[H]
    \caption{Statistics of the datasets used in the experiments. $|L|$ denotes the number of labels,  Avg($|L|$) is the average number of labels per instance and Max($|L|$) is the maximum number of labels for an instance.}
    \label{tab:data}
    \begin{adjustbox}{max width=0.48\textwidth}
    \begin{tabular}{cccccccc} 
    \toprule
    \textbf{Dataset} & \textbf{Hierarchy} & \textbf{\textbar{}L\textbar{}} & \textbf{Avg(\textbar{}L\textbar{})} & \textbf{Max(\textbar{}L\textbar{})} & \textbf{Train} & \textbf{Val} & \textbf{Test} \\ 
    \hline
    RCV1 & Tree & 104 & 3.24 & 17 & 20833 & 2314 & 781265 \\
    NYT & Tree & 120 & 6.58 & 24 & 86461 & 9606 & 9903 \\
    Yelp & DAG & 539 & 4.07 & 32 & 98460 & 10939 & 46884 \\
    \bottomrule
    \end{tabular}
    \end{adjustbox}
\end{table}

\subsection{Comparison of models that do not use the true hierarchy}
We compare performance of the different models that do not use the true hierarchy. These include our flat baseline \modelflt, the cascaded model \modelcas\ as well as our joint model \modeljnt. We compare them against the baseline TextCNN-flat model reported in~\citet{emnlp}. The results are presented in Table~\ref{tab:main} for $\lambda=0.1$.

Overall, our baseline (\modelflt) performs better than the previous baseline with the exception of Macro-F1 on NYT. We observe improvement of the joint model \modeljnt\ over the flat (\modelflt) and cascaded (\modelcas) models on RCV1 and NYT (for each of which, the labels form a tree) in Table~\ref{tab:main}. However, on the Yelp dataset, the cascaded model (\modelcas) performs somewhat worse (-2 Micro-F1 and -3.3 Macro-F1) than our baseline model (\modelflt), hinting at the possibility that label co-occurrence information might not be helpful toward the classification task. This could be partly also because the labels in Yelp are structured in the form of a DAG. Constant curvature property of hyperbolic spaces makes them unsuitable for learning DAG structures \cite{li2018smoothing}. However, the Macro-F1 performance of \modeljnt\ is far better than that of the cascaded model \modelcas. This illustrates that our joint model is able to better recover from less reliable (or less useful) label co-occurrence information, just as was illustrated in the Table~\ref{tab:synth1} for the synthetic setting.

\begin{table}[!ht]
\centering
\caption{Performance comparison on all three datasets with TextCNN as the base classification model. Recall that \modelflt\ is our own multi-class classification baseline.
We observe that the numbers reported by~\citet{emnlp} (indicated by $^*$) for the TextCNN based flat baseline model are consistently outperformed by the proposed \model\ models.}
\label{tab:main}
\arrayrulecolor[rgb]{0.192,0.192,0.192}
\begin{adjustbox}{max width=0.48\textwidth}
\begin{tabular}{llll} 
\toprule
\multicolumn{1}{l}{Dataset} & Method & Micro-F1 & Macro-F1 \\ 
\toprule
& TextCNN-Flat$^*$ & 76.6 & 43.0 \\
&\modelflt & 77.9 & 44.5 \\
\multicolumn{1}{c}{RCV1}&\modelcas & 78.0 & 45.5 \\
 & \modeljnt & \textbf{79.3} & \textbf{47.3} \\
\hline\hline
& TextCNN-Flat$^*$ & 69.5 & 39.5 \\
&\modelflt & 76.4 & 37.1 \\
\multicolumn{1}{c}{NYTimes}& \modelcas & 74.6 & 33.2 \\ 
 & {\modeljnt} & {\textbf{77.0}} & {\textbf{43.6}} \\

\hline\hline
& TextCNN-Flat$^*$ & \textbf{62.8} & 27.3 \\
&\modelflt & 62.5& \textbf{37.9}\\
\multicolumn{1}{c}{Yelp} & \modelcas  & 60.5 & 33.9\\
 & {\modeljnt } & 60.8 & 35.6 \\

\bottomrule
\end{tabular}
\end{adjustbox}
\arrayrulecolor{black}
\end{table}

\subsection{Comparison of Hyperbolic space and Euclidean space}
To assess the utility of the hyperbolic space for embedding hierarchical labels, we compare \modeljnt\ and \modeleuc. Table~\ref{tab:main-eucjnt} presents this comparison on the three datasets. \modeleuc\ performs worse than \modeljnt\ which uses the hyperbolic space for embedding labels (except Micro-F1 for Yelp due to reasons stated before). This is expected since embedding trees is much more effective in the hyperbolic space compared to Euclidean space since in the hyperbolic space, volume grows exponentially with distance from the origin while in Euclidean space, this growth is polynomial. The number of nodes in a tree also increases exponentially with distance from the root, making Hyperbolic spaces useful for embedding hierarchies.

\begin{table}[!h]
\centering
\caption{Performance comparison for \modeljnt\ with \modeleuc. \modeljnt\ consistently has better Macro-F1 better than \modeleuc\ and generally better Micro-F1 too. This illustrates the utility of Hyperbolic spaces for embeddding label hierarchies}
\label{tab:main-eucjnt}
\arrayrulecolor[rgb]{0.192,0.192,0.192}
\begin{adjustbox}{max width=0.48\textwidth}
\begin{tabular}{llll} 
\toprule
\multicolumn{1}{l}{Dataset} & Method & Micro-F1 & Macro-F1 \\ 
\toprule
& \modeleuc & 78.4 & 47.6 \\
\multicolumn{1}{c}{RCV1}
& \modeljnt & \textbf{79.3} & \textbf{47.3} \\
\hline\hline
& \modeleuc & 76.4 & 40.4 \\
\multicolumn{1}{c}{NYTimes}
 & {\modeljnt}& \textbf{77.0} & \textbf{43.6} \\
\hline\hline
& \modeleuc & \textbf{61.1}& 34.2\\
\multicolumn{1}{c}{Yelp} 
& {\modeljnt} & 60.8 & \textbf{35.6} \\
\bottomrule
\end{tabular}
\end{adjustbox}
\arrayrulecolor{black}
\end{table}

\subsection{Comparison with model that explicitly uses the true hierarchy}
We compare performance of our joint approach \modeljnt\ against a state-of-the-art hierarchical multi-label classifier, HiLAP \citep{emnlp}. However, unlike our proposed models (variants of \model), HiLAP has access to the true hierarchy both training and inference. Thus, HiLAP serves as some form of skyline for the \model\ suite of approaches proposed in this paper. HiLAP learns label assignment policy using the reinforcement learning framework. 

In Table \ref{tab:hilap}, we compare the performance of \modeljnt\ against HiLAP model as reported in \citet{emnlp}. Interestingly, on RCV1, we obtain better Micro-F1 score ($+0.7$) for the joint model \modeljnt\ over the HiLAP method. On NYT, our Micro-F1 score is far better ($+7.1$) than HiLAP; our Macro-F1 score is also marginally better ($+0.4$) than their Macro-F1 scores. These results are interesting because \modeljnt\ seems to obtain better generalisation through joint learning of the document classifier and label embeddings in a hyperbolic space, even without access to the true hierarchy. However, on Yelp, HiLAP seems to benefit over \modeljnt\ by explicitly using the true hierarchy.

\begin{table}[!h]
\centering
\caption{Performance comparison of \modeljnt\ with HiLAP with respect to Macro-F1 and Micro-F1. It it interesting to note here that though \modeljnt\ does not know the true hierarchy, it performs better than HiLAP (which uses the true hierarchy) in some cases. (HiLAP numbers are those reported by \citet{emnlp})}
\label{tab:hilap}
\begin{adjustbox}{max width=0.48\textwidth}
\begin{tabular}{ccclcc} 
\toprule
\multicolumn{1}{l}{Dataset} & \multicolumn{2}{c}{\modeljnt} &  & \multicolumn{2}{c}{HiLAP} \\ 
\cline{2-3}\cline{5-6}
\multicolumn{1}{l}{} & \multicolumn{1}{l}{Micro} & \multicolumn{1}{l}{Macro} &  & Micro & Macro \\ 
\toprule
RCV1 & \textbf{79.3} & 47.3 &  & 78.6 & \textbf{50.5} \\
NYTimes & \textbf{77.0} & \textbf{43.6} &  & 69.9 & 43.2 \\
Yelp & 60.8 & 35.6 &  & \textbf{65.5} & \textbf{37.3} \\
\bottomrule
\end{tabular}
\end{adjustbox}
\end{table}

\subsection{Evaluating performance of embeddings}
\begin{figure}[!h]
    \centering
    \includegraphics[width=.4\textwidth]{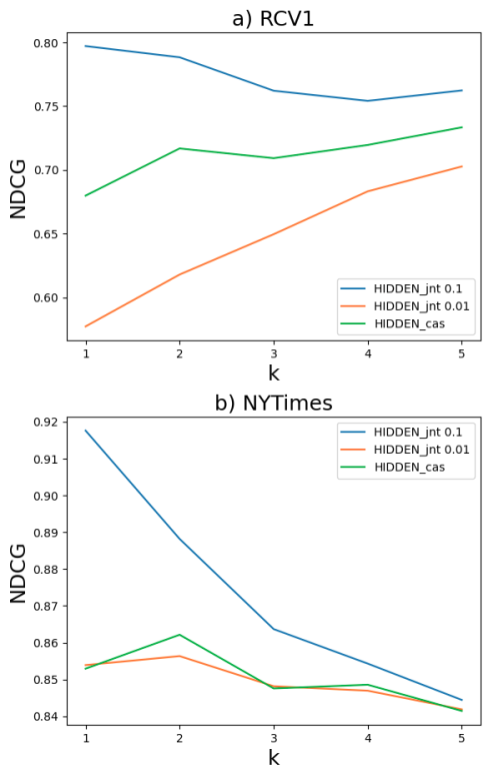}
    \caption{Plot of NDCG versus k for assessing the quality of the learnt label embeddings with respect to the actual hierarchy on RCV1 and NYT datasets. The better performance of \modeljnt\ indicates the label embeddings $\Theta_{\text{jnt}}$ are most representative of the true hierarchy.}
    \label{fig:embed}
\end{figure}

We compare embeddings learned using different approaches with the ground truth hierarchy to evaluate the effectiveness of the embeddings. Figure \ref{fig:embed} shows the plot of NDCG scores for different values of k on the RCV1 and NYTimes dataset across \modelcas\ and \modeljnt\ (for two different values of $\lambda$). In Table~\ref{tab:correlation}, we compare the Spearman rank correlation. The superior performance of \modeljnt strongly suggests that the embeddings learnt using the joint model are more representative of the true hierarchical organisation of the labels than those obtained using the flat and cascaded variants. This also goes to show that even the first term in our objective has positive contribution towards the learning of hyperbolic embeddings and indeed joint learning is beneficial.

\begin{table}[!h]
\centering
\caption{Spearman rank correlation test for the generated embeddings for all the datasets. Each method is compared against the ground truth hierarchy. Highest values for \modeljnt\ indicates the label embeddings $\Theta_{\text{jnt}}$ are most representative of the true hierarchy.}
\label{tab:correlation}
\begin{adjustbox}{max width=0.48\textwidth}
\begin{tabular}{cccc} 
\toprule
& \modelflt & \modeljnt & \modelcas \\
\toprule
RCV1 & 21.2 & \textbf{53.9} & 44.1 \\
NYTimes & 11.4 & \textbf{39.5}   & 36.1 \\
Yelp & 16.3 & \textbf{31.9}   & 28.8 \\
\bottomrule
\end{tabular}
\end{adjustbox}
\end{table}

\section{Conclusion}
We propose a novel approach to hierarchical multi-label classification based on joint learning of document classifier and label embeddings in hyperbolic space. The proposed framework \model\ allows us to discover label hierarchical relationship by leveraging properties of hyperbolic geometry. Even though label-hierarchy is assumed to be unavailable, our method achieves comparable results with state-of-the-art \textit{hierarchy aware} methods. We performed extensive experiments on three datasets and demonstrate effectiveness of the learned embeddings.

\bibliography{anthology,eacl2021}

\begin{thebibliography}{32}
\expandafter\ifx\csname natexlab\endcsname\relax\def\natexlab#1{#1}\fi

\bibitem[{Agrawal et~al.(2013)Agrawal, Gupta, Prabhu, and
  Varma}]{agrawal2013multi}
Rahul Agrawal, Archit Gupta, Yashoteja Prabhu, and Manik Varma. 2013.
\newblock Multi-label learning with millions of labels: Recommending advertiser
  bid phrases for web pages.
\newblock In \emph{Proceedings of the 22nd international conference on World
  Wide Web}, pages 13--24.

\bibitem[{Bhatia et~al.(2016)Bhatia, Dahiya, Jain, Mittal, Prabhu, and
  Varma}]{extreme}
Kush Bhatia, Kunal Dahiya, Himanshu Jain, Anshul Mittal, Yashoteja Prabhu, and
  Manik Varma. 2016.
\newblock \href {http://manikvarma.org/downloads/XC/XMLRepository.html} {The
  extreme classification repository: Multi-label datasets and code}.

\bibitem[{Cerri et~al.(2011)Cerri, Barros, and
  de~Carvalho}]{cerri2011hierarchical}
Ricardo Cerri, Rodrigo~C Barros, and Andr{\'e}~CPLF de~Carvalho. 2011.
\newblock Hierarchical multi-label classification for protein function
  prediction: A local approach based on neural networks.
\newblock In \emph{2011 11th International Conference on Intelligent Systems
  Design and Applications}, pages 337--343. IEEE.

\bibitem[{Cesa-Bianchi et~al.(2006)Cesa-Bianchi, Gentile, and
  Zaniboni}]{cesa2006hierarchical}
Nicol{\`o} Cesa-Bianchi, Claudio Gentile, and Luca Zaniboni. 2006.
\newblock Hierarchical classification: combining bayes with svm.
\newblock In \emph{Proceedings of the 23rd international conference on Machine
  learning}, pages 177--184.

\bibitem[{Chen et~al.(2019)Chen, Huang, Xiao, Cai, and Jing}]{hyperim}
Boli Chen, Xin Huang, Lin Xiao, Zixin Cai, and Liping Jing. 2019.
\newblock Hyperbolic interaction model for hierarchical multi-label
  classification.
\newblock \emph{arXiv preprint arXiv:1905.10802}.

\bibitem[{Gopal and Yang(2013)}]{gopal2013recursive}
Siddharth Gopal and Yiming Yang. 2013.
\newblock Recursive regularization for large-scale classification with
  hierarchical and graphical dependencies.
\newblock In \emph{Proceedings of the 19th ACM SIGKDD international conference
  on Knowledge discovery and data mining}, pages 257--265.

\bibitem[{Gromov(1987)}]{hyperbolic1987}
Mikhael Gromov. 1987.
\newblock Hyperbolic groups.
\newblock In \emph{Essays in group theory}, pages 75--263. Springer.

\bibitem[{J{\"a}rvelin and Kek{\"a}l{\"a}inen(2002)}]{dcg}
Kalervo J{\"a}rvelin and Jaana Kek{\"a}l{\"a}inen. 2002.
\newblock Cumulated gain-based evaluation of ir techniques.
\newblock \emph{ACM Transactions on Information Systems (TOIS)},
  20(4):422--446.

\bibitem[{Johnson and Zhang(2015)}]{johnson2015effective}
Rie Johnson and Tong Zhang. 2015.
\newblock Effective use of word order for text categorization with
  convolutional neural networks.
\newblock In \emph{Proceedings of the 2015 Conference of the North American
  Chapter of the Association for Computational Linguistics: Human Language
  Technologies}, pages 103--112.

\bibitem[{Kim(2014)}]{textcnn}
Yoon Kim. 2014.
\newblock Convolutional neural networks for sentence classification.
\newblock In \emph{Proceedings of the 2014 Conference on Empirical Methods in
  Natural Language Processing (EMNLP)}, pages 1746--1751.

\bibitem[{Kingma and Ba(2014)}]{kingma2014adam}
Diederik~P Kingma and Jimmy Ba. 2014.
\newblock Adam: A method for stochastic optimization.
\newblock \emph{arXiv preprint arXiv:1412.6980}.

\bibitem[{Koller and Sahami(1997)}]{koller1997hierarchically}
Daphne Koller and Mehran Sahami. 1997.
\newblock Hierarchically classifying documents using very few words.
\newblock Technical report, Stanford InfoLab.

\bibitem[{Krioukov et~al.(2010)Krioukov, Papadopoulos, Kitsak, Vahdat, and
  Bogun{\'a}}]{complexhyperbolic}
Dmitri Krioukov, Fragkiskos Papadopoulos, Maksim Kitsak, Amin Vahdat, and
  Mari{\'a}n Bogun{\'a}. 2010.
\newblock Hyperbolic geometry of complex networks.
\newblock \emph{Physical Review E}, 82(3):036106.

\bibitem[{Lewis et~al.(2004)Lewis, Yang, Rose, and Li}]{rcv1}
David~D Lewis, Yiming Yang, Tony~G Rose, and Fan Li. 2004.
\newblock Rcv1: A new benchmark collection for text categorization research.
\newblock \emph{Journal of machine learning research}, 5(Apr):361--397.

\bibitem[{Li et~al.(2018)Li, Vilnis, Zhang, Boratko, and
  McCallum}]{li2018smoothing}
Xiang Li, Luke Vilnis, Dongxu Zhang, Michael Boratko, and Andrew McCallum.
  2018.
\newblock Smoothing the geometry of probabilistic box embeddings.
\newblock In \emph{International Conference on Learning Representations}.

\bibitem[{Liu et~al.(2017)Liu, Chang, Wu, and Yang}]{liu2017deep}
Jingzhou Liu, Wei-Cheng Chang, Yuexin Wu, and Yiming Yang. 2017.
\newblock Deep learning for extreme multi-label text classification.
\newblock In \emph{Proceedings of the 40th International ACM SIGIR Conference
  on Research and Development in Information Retrieval}, pages 115--124.

\bibitem[{Mao et~al.(2019)Mao, Tian, Han, and Ren}]{emnlp}
Yuning Mao, Jingjing Tian, Jiawei Han, and Xiang Ren. 2019.
\newblock Hierarchical text classification with reinforced label assignment.
\newblock In \emph{Proceedings of the 2019 Conference on Empirical Methods in
  Natural Language Processing and the 9th International Joint Conference on
  Natural Language Processing (EMNLP-IJCNLP)}, pages 445--455.

\bibitem[{McCallum(1999)}]{mccallum1999multi}
Andrew~Kachites McCallum. 1999.
\newblock Multi-label text classification with a mixture model trained by em.
\newblock In \emph{AAAI 99 workshop on text learning}. Citeseer.

\bibitem[{Nickel and Kiela(2017)}]{poincare}
Maximillian Nickel and Douwe Kiela. 2017.
\newblock Poincar{\'e} embeddings for learning hierarchical representations.
\newblock In \emph{Advances in neural information processing systems}, pages
  6338--6347.

\bibitem[{Nickel and Kiela(2018)}]{lorentz}
Maximillian Nickel and Douwe Kiela. 2018.
\newblock Learning continuous hierarchies in the lorentz model of hyperbolic
  geometry.
\newblock In \emph{International Conference on Machine Learning}, pages
  3779--3788.

\bibitem[{Peng et~al.(2018)Peng, Li, He, Liu, Bao, Wang, Song, and
  Yang}]{peng2018large}
Hao Peng, Jianxin Li, Yu~He, Yaopeng Liu, Mengjiao Bao, Lihong Wang, Yangqiu
  Song, and Qiang Yang. 2018.
\newblock Large-scale hierarchical text classification with recursively
  regularized deep graph-cnn.
\newblock In \emph{Proceedings of the 2018 World Wide Web Conference}, pages
  1063--1072.

\bibitem[{Peng et~al.(2016)Peng, You, Wang, Zhai, Mamitsuka, and
  Zhu}]{peng2016deepmesh}
Shengwen Peng, Ronghui You, Hongning Wang, Chengxiang Zhai, Hiroshi Mamitsuka,
  and Shanfeng Zhu. 2016.
\newblock Deepmesh: deep semantic representation for improving large-scale mesh
  indexing.
\newblock \emph{Bioinformatics}, 32(12):i70--i79.

\bibitem[{Pennington et~al.(2014)Pennington, Socher, and Manning}]{glove}
Jeffrey Pennington, Richard Socher, and Christopher~D Manning. 2014.
\newblock Glove: Global vectors for word representation.
\newblock In \emph{Proceedings of the 2014 conference on empirical methods in
  natural language processing (EMNLP)}, pages 1532--1543.

\bibitem[{Rousu et~al.(2006)Rousu, Saunders, Szedmak, and
  Shawe-Taylor}]{rousu2006kernel}
Juho Rousu, Craig Saunders, Sandor Szedmak, and John Shawe-Taylor. 2006.
\newblock Kernel-based learning of hierarchical multilabel classification
  models.
\newblock \emph{Journal of Machine Learning Research}, 7(Jul):1601--1626.

\bibitem[{Sandhaus(2008)}]{nytimes}
Evan Sandhaus. 2008.
\newblock The new york times annotated corpus.
\newblock \emph{Linguistic Data Consortium, Philadelphia}, 6(12):e26752.

\bibitem[{Silla and Freitas(2011)}]{survey}
Carlos~N Silla and Alex~A Freitas. 2011.
\newblock A survey of hierarchical classification across different application
  domains.
\newblock \emph{Data Mining and Knowledge Discovery}, 22(1-2):31--72.

\bibitem[{Silla~Jr and Freitas(2009)}]{silla2009global}
Carlos~N Silla~Jr and Alex~A Freitas. 2009.
\newblock A global-model naive bayes approach to the hierarchical prediction of
  protein functions.
\newblock In \emph{2009 Ninth IEEE International Conference on Data Mining},
  pages 992--997. IEEE.

\bibitem[{Triguero and Vens(2016)}]{triguero2016labelling}
Isaac Triguero and Celine Vens. 2016.
\newblock Labelling strategies for hierarchical multi-label classification
  techniques.
\newblock \emph{Pattern Recognition}, 56:170--183.

\bibitem[{Wang et~al.(2001)Wang, Zhou, and He}]{wang2001hierarchical}
Ke~Wang, Senqiang Zhou, and Yu~He. 2001.
\newblock Hierarchical classification of real life documents.
\newblock In \emph{Proceedings of the 2001 SIAM International Conference on
  Data Mining}, pages 1--16. SIAM.

\bibitem[{Wehrmann et~al.(2018)Wehrmann, Cerri, and Barros}]{hmcn}
Jonatas Wehrmann, Ricardo Cerri, and Rodrigo Barros. 2018.
\newblock Hierarchical multi-label classification networks.
\newblock In \emph{International Conference on Machine Learning}, pages
  5075--5084.

\bibitem[{Yang et~al.(2009)Yang, Sun, Wang, and Chen}]{yang2009effective}
Bishan Yang, Jian-Tao Sun, Tengjiao Wang, and Zheng Chen. 2009.
\newblock Effective multi-label active learning for text classification.
\newblock In \emph{Proceedings of the 15th ACM SIGKDD international conference
  on Knowledge discovery and data mining}, pages 917--926.

\bibitem[{Zar(2005)}]{zar2005spearman}
Jerrold~H Zar. 2005.
\newblock Spearman rank correlation.
\newblock \emph{Encyclopedia of Biostatistics}, 7.

\end{thebibliography}
\bibliographystyle{acl_natbib}

\newpage
\appendix

\section{Explanation for $\Pi(x)$}
The Lorentz model is defined as the Riemannian Manifold, $\mathcal{L}^{n} = (\mathcal{H}^n, g_l)$, where $\mathcal{H}^n = \{ \mathbf{x} \in \mathbb{R}^{n+1} :\ \langle \mathbf{x},\mathbf{x}\rangle_\mathcal{L} = -1, \mathbf{x}_0 > 0\}$, and $g_l=diag([-1\ 1\ \ldots\ 1])$. Here, $\langle \mathbf{x},\mathbf{y}\rangle{_\mathcal{L}}$, known as the Minkowski inner-product, is given by 
$$\langle \mathbf{x},\mathbf{y}\rangle{_\mathcal{L}} = -x_0 y_0 + x_1 y_1 + \ldots + x_{n} y_{n}$$

The Poincar\'e model and the Lorentz model are equivalent in isometry. Therefore points in Lorentz manifold can be mapped into Poincar\'e ball as, $p: \mathcal{H}^n \rightarrow \mathcal{P}^n $
$$p(x_0, x_1,\ldots, x_n) = \frac{(x_1, \ldots, x_n)}{x_0 + 1}$$

A point $\mathbf{x}$ in the Euclidean space $\mathbb{R}^n$ can be projected onto the Lorentz manifold $\mathcal{H}^n$ using the transformation $\Omega(\mathbf{x)} = \left[\sqrt{1+\|\mathbf{x}\|^2}, \mathbf{x}\right]$. This transformation ensures that the Minkowski inner-product, $\left\langle\Omega(\mathbf{x}),\Omega(\mathbf{x})\right\rangle_{\mathcal{L}},$ is equal to $-1$, and that the first component, $\Omega(\mathbf{x})_0\equiv\sqrt{1+\|\mathbf{x}\|^2}$ is positive, as required for membership in the Lorentz manifold. 

Now using the isometry between Poincar\'e and Lorentz models \citep{lorentz}, we have $\Pi: \mathbb{R}^n \rightarrow \mathcal{P}^n$ as 
$$\Pi(\mathbf{x}) = p\left(\Omega(\mathbf{x})\right) = \frac{\mathbf{x}}{1+\sqrt{1+\|\mathbf{x}\|^2}}$$

\section{Dataset Details}
We describe the details of the datasets used in our experiments.

For RCV1 dataset \citep{rcv1}, we use the original training/test split and use 10\% of the training set as the validation set. We introduce an extra \textit{Root} label in addition to the 103 labels present in the dataset. Each document in the dataset is labelled with this label.

The details for the other datasets used are same as in \citet{emnlp} and we refer the readers to the same.

\end{document}